\documentclass[10pt,twocolumn,letterpaper]{article}

\usepackage{iccv}
\usepackage{times}
\usepackage{epsfig}
\usepackage{graphicx}
\usepackage{caption}
\usepackage{subcaption}
\usepackage{amsmath}
\usepackage{amssymb}

\DeclareMathOperator*{\argmin}{arg\,min}

\usepackage[pagebackref=true,breaklinks=true,letterpaper=true,colorlinks,bookmarks=false]{hyperref}

\iccvfinalcopy 

\ificcvfinal\fi
\begin{document}

\title{Learning Local RGB-to-CAD Correspondences for Object Pose Estimation}

\author{Georgios Georgakis$^{1}$, Srikrishna Karanam$^{2}$, Ziyan Wu$^{2}$, and  Jana Ko{\v{s}}eck{\'a}$^{1}$\\
$^{1}$Department of Computer Science, George Mason University, Fairfax VA\\
$^{2}$Siemens Corporate Technology, Princeton NJ\\
{\tt\small ggeorgak@gmu.edu,\{first.last\}@siemens.com,kosecka@cs.gmu.edu}
}


\twocolumn[{%
\renewcommand\twocolumn[1][]{#1}%
\maketitle
\vspace{-2em}
\begin{center}
    \centering
    \includegraphics[scale=0.7]{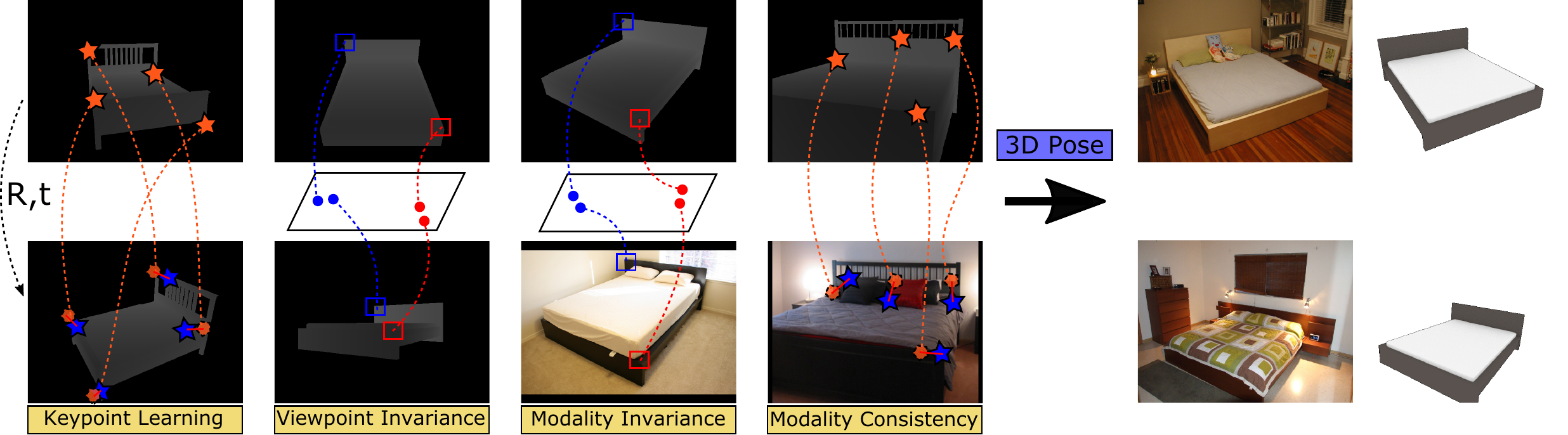}
    \captionof{figure}{We present a new method that matches RGB images to rendered depth images of CAD models for object pose estimation. The method does not require either textured CAD models or 3D pose annotations for RGB images during training. This is achieved through a series of constraints that enforce viewpoint and modality invariance for local features, and learn how to select keypoints consistently across modalities.}
    \label{fig:title}
\end{center}%
}]

\begin{abstract}
\vspace*{-\parskip}

We consider the problem of 3D object pose estimation. While much recent work has focused on the RGB domain, the reliance on accurately annotated images limits their generalizability and scalability. On the other hand, the easily available CAD models of objects are rich sources of data, providing a large number of synthetically rendered images. In this paper, we solve this key problem of existing methods requiring expensive 3D pose annotations by proposing a new method that matches RGB images to CAD models for object pose estimation. Our key innovations compared to existing work include removing the need for either real-world textures for CAD models or explicit 3D pose annotations for RGB images. We achieve this through a series of objectives that learn how to select keypoints and enforce viewpoint and modality invariance across RGB images and CAD model renderings. We conduct extensive experiments to demonstrate that the proposed method can reliably estimate object pose in RGB images, as well as generalize to object instances not seen during training. 

\end{abstract}

\section{Introduction}

Estimating the 3D pose of objects is an important capability for enabling robots' interaction with real environments and objects as well as augmented reality applications. While several approaches to this problem assume RGB-D data, most mobile and wearable cameras are not paired with a depth sensor, prompting recent research focus on the RGB domain. Furthermore, even though several methods have shown promising results on 3D object pose estimation with real RGB images, they either require accurate 3D annotations or 3D object models with realistic textures in the training stage. Currently available datasets are not large enough to capture real world diversity, limiting the potential of these methods in generalizing to a variety of applications. In addition, capturing real RGB data and manual pose annotation is an arduous procedure. 

The problem of object pose estimation is an inherently 3D problem; it is the shape of the object which gives away its pose regardless of its appearance. Instead of attempting to learn an intrinsic decomposition of images~\cite{tenenbaum2017intrinsic}, we focus on finding the association of parts of objects depicted in RGB images with their counterparts in 3D depth images. Ideally, we would like to learn this association in order to establish correspondences between a query RGB image and a rendered depth image from a CAD model, without requiring any existing 3D annotations. This, however, requires us to address the problem of the large appearance gap between these two modalities.

In this paper, we propose a new framework for estimating the 3D pose of objects in RGB images, using only 3D textureless CAD models of objects instances. The easily available CAD models can generate a large number of synthetically rendered depth images from multiple viewpoints. In order to address the aforementioned problems, we define a {\em quadruplet} convolutional neural network to jointly learn keypoints and their associated descriptors for robust matching between different modalities and changes in viewpoint. The general idea is to learn the keypoint locations using a pair of rendered depth images from a CAD model from two different poses, followed by learning how to match keypoints across modalities using an aligned RGB-D image pair. Figure~\ref{fig:title} outlines our training constraints.
At test time, given a query RGB image, we extract keypoints and their representations and match them with a database of keypoints and their associated descriptors extracted from rendered depth images. These are used to establish 2D-3D correspondences, followed by a RANSAC and PnP algorithm for pose estimation. 

To summarize, the key contributions of this work are the following:
\begin{itemize}
\item We present a new framework for 3D object pose estimation using only textureless CAD models and aligned RGB-D frames in the training stage, without explicitly requiring 3D pose annotations for the RGB images. 
\item We present an end-to-end learning approach for keypoint selection optimized for the relative pose estimation objective, and transfer of keypoint predictions and their representations from rendered depth to RGB images.
\item We demonstrate the generalization capability of our method to new (unseen during training) instances of the same object category. 
\end{itemize}

\section{Related Work}

There is a large body of work on 3D object pose estimation. Here, we review existing methods based on the type and the amount of used training data and its modalities.

\noindent
{\bf Using 3D textured instance models.} Notable effort was devoted to the problem of pose estimation for object instances from images, where 3D textured instance models were available during the training stage~\cite{hinterstoisser2011multimodal, collet2011moped, tang2012textured}. Early isolated approaches led to the development of more recent benchmarks for this problem~\cite{6DposebenchmarkECCV2018}. Traditional approaches of this type included template matching~\cite{hinterstoisser2011multimodal, zhu2014single}, where the target pose is retrieved from the best matched model in a database, and local descriptor matching~\cite{collet2011moped, tang2012textured}, where hand-engineered descriptors such as SIFT~\cite{lowe2004distinctive} are used to establish 2D-3D correspondences with a 3D object model followed by the PnP algorithm for 6-DoF pose.
Additionally, some works employed a patch-based dense voting scheme~\cite{brachmann2014learning, tejani2014latent, doumanoglou2016recovering, kehl2016deep}, where a function is learned to map local representations to 3D coordinates or to pose space. However, these approaches assume that the 3D object models were created from real images and contain realistic textures. In contrast, our work uses only textureless CAD models of object instances. \\ 
\noindent
{\bf 2D-to-3D alignment with CAD models.}
Other work has sought to solve 3D object pose estimation as a 2D-to-3D alignment problem by utilizing object CAD models~\cite{aubry2014seeing, massa2016deep, lim2013parsing, izadinia2017im2cad, bansal2016marr, rad2018domain}. For example, Aubry \etal~\cite{aubry2014seeing} learned part-based exemplar classifiers from textured CAD models and applied them on real images to establish 2D-3D correspondences. In a similar fashion, Lim \etal~\cite{lim2013parsing} trained a patch detector from edge maps for each interest point. The work of Massa \etal~\cite{massa2016deep} learned how to match view-dependent exemplar features by adapting the representations extracted from real images to their CAD model counterparts. 
The closest work to ours in this area is Rad \etal~\cite{rad2018domain}, which attempts to bridge the domain gap between real and synthetic depth images, by learning to map color features to real depth features and subsequently to synthetic depth features.
In their attempt to bridge the gap between the two modalities, these approaches were required to either learn a huge number of exemplar classifiers, or learn how to adapt features for each specific category and viewpoint. We avoid this problem by simply adapting keypoint predictions and descriptors between the two modalities. \\
\noindent
{\bf Pose estimation paired with object detection.}
With the recent success of deep convolutional neural networks (CNN) on object recognition and detection, many works extended 3D object instance pose estimation to object categories,  from an input RGB image~\cite{mahendran20173d, mottaghi2015coarse, mousavian20173d, xiang2017posecnn, poirson2016fast, kehl2017ssd, liu2016ssd, kundu20183d}. In Mahendran \etal \cite{mahendran20173d} a 3D pose regressor was learned for each object category.  In Mousavian \etal \cite{mousavian20173d}, a discrete-continuous formulation for the pose prediction was introduced, which first classified the orientation to a discrete set of bins and then regressed the exact angle within the bin. Poirson \etal \cite{poirson2016fast} and Kehl \etal \cite{kehl2017ssd} both extended the SSD~\cite{liu2016ssd} object detector to predict azimuth and elevation or the 6-DoF pose respectively.
In Kundu \etal \cite{kundu20183d}, an analysis-by-synthesis approach was introduced, in which, given predicted pose and shape, the object was rendered and compared to 2D instance segmentation annotations. 
All of these approaches require 3D pose annotations for the RGB images during training,
as opposed to our work, which only needs the CAD models of the objects.

\noindent
{\bf Keypoint-based methods.}
Another popular direction in the pose estimation literature is learning how to estimate 
keypoints, which can be used to infer the pose. These methods are usually motivated by the presence of occlusions~\cite{pavlakos20176, hueting2017seethrough} and require keypoint annotations.
For example, Wu \etal \cite{wu20183d} trained a model for 2D keypoint prediction on real images and
estimated the 3D wireframes of objects using a model trained on synthetic shapes. The 3D wireframe is then projected to real images labeled with 2D keypoints to enforce consistency. In Li \etal \cite{li2017deep}, the authors manually annotated 3D keypoints on textured CAD models and generated a synthetic dataset which provides multiple layers of supervision during training, while Tekin \etal \cite{tekin2018real} learned to predict the 2D image locations of the projected vertices of an object's 3D bounding box before using the PnP algorithm for pose estimation. Furthermore, Tulsiani \etal \cite{tulsiani2015viewpoints} exploited the relationship between viewpoint and visible keypoints and refined an existing coarse pose estimation using keypoint predictions. 
Our work, rather than relying on existing keypoint annotations, optimizes the keypoint selection based on a relative pose estimation objective. Related approaches also learn keypoints ~\cite{suwajanakorn2018discovery, georgakis2018end, yi2016lift, zhou2018unsupervised},  but either rely on hand-crafted detectors to collect training data~\cite{yi2016lift}, or do not extend to real RGB pose estimation~\cite{suwajanakorn2018discovery, georgakis2018end, zhou2018unsupervised}.

\noindent
{\bf Synthetic data generation.}
In an attempt to address the scarcity of annotated data, some approaches rely on the generation of large amounts of synthetic data for training~\cite{sun2018pix3d, su2015render, gupta2015inferring}. A common technique is to render textured CAD models and superimpose them on real backgrounds. In order to ensure diversity in the training data, rendering parameters such as pose, shape deformations, and illumination are randomly chosen. However, training exclusively on synthetic data has shown to be detrimental to the learned representations as the underlying statistics of real RGB images are usually very different.

\begin{figure}[t]
\begin{center}
\includegraphics[width=1\linewidth]{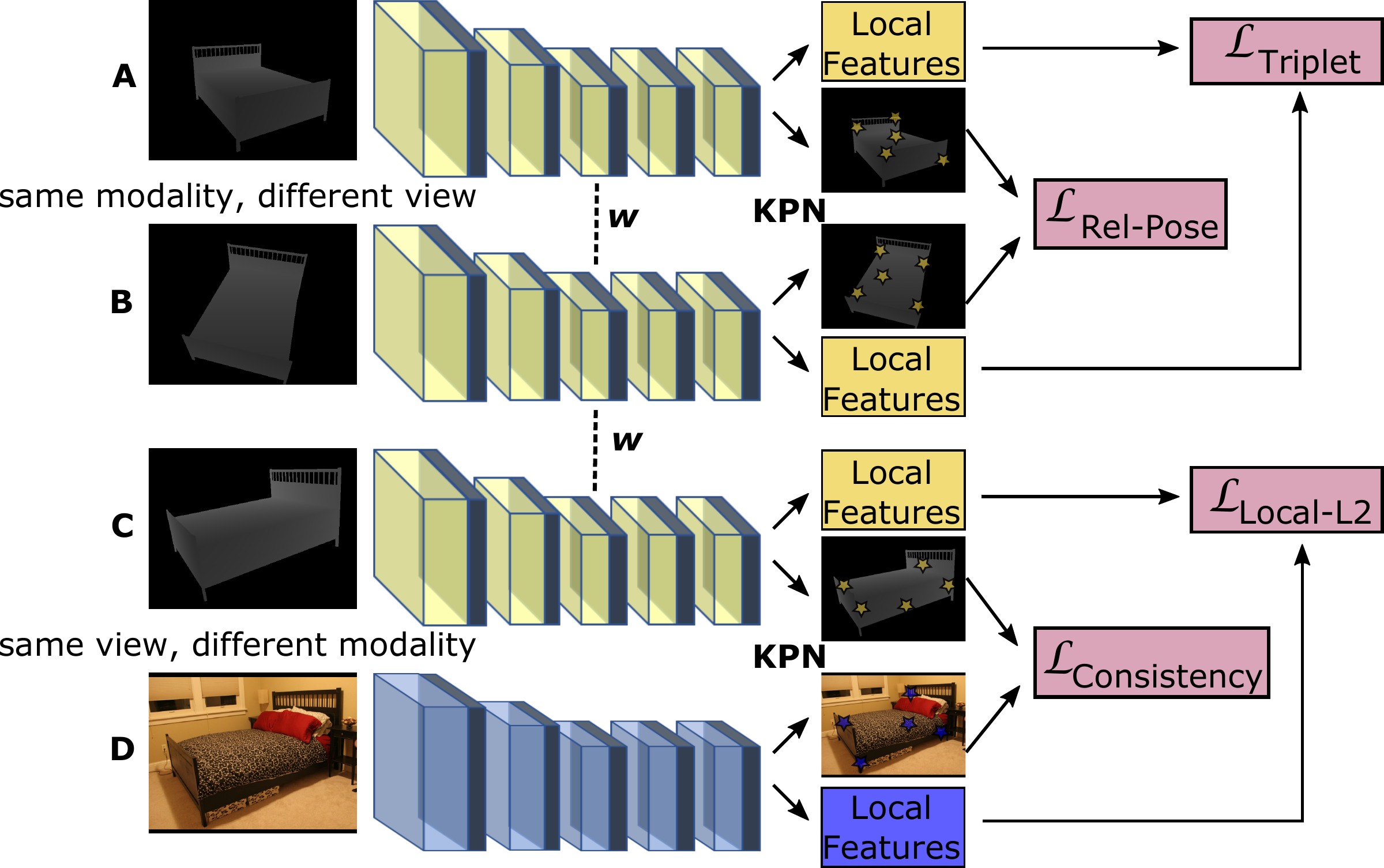}
\end{center}
   \caption{Outline of the proposed architecture depicting the four branches of the network, their inputs, and the objectives imposed during training. The color coding of the CNNs signifies weight sharing.}
\vspace{-0.5cm}
\label{fig:overview}
\end{figure}

\section{Approach}
We are interested in estimating the 3D pose of objects in RGB images by matching keypoints to the object's CAD model. Our work does not make use of pose annotations, but instead relies on CAD model renderings of different poses that are easily obtained with an off-the-shelf renderer, such as Blender~\cite{blender}. These rendered depth images are used to learn keypoints and their representations optimized for the task of pose estimation.  
The learned representations are then transferred to the RGB domain. In summary, our work can be divided into four objectives: keypoint learning, view-invariant descriptors, modality-invariant descriptors, and modality consistent keypoints. 

Specifically, each training input is provided as a quadruplet of images, consisting of a pair of rendered depth images sampled from the object's view sphere and a pair of aligned depth and RGB images (see Figure~\ref{fig:overview}). For each image, we predict a set of keypoints and their local representations, but the optimization objectives differ for the various branches. For the first two branches A and B, $L_{rel\_pose}$ loss enforces the pose consistency of the keypoints selection and the similarity of keypoint descriptors for their matching is enforced using a triplet loss $L_{triplet}$. The two bottom branches C and D are utilized to enforce consistent keypoint prediction between the depth and the RGB modalities $L_{consistency}$ and for matching their local representations across the modalities $L_{local\_l2}$. The general idea of our approach is to learn informative keypoints and their associated local descriptors from abundant rendered depth images and transfer this knowledge to the RGB data.

\noindent
{\bf Architecture.}
Our proposed architecture is a Quadruplet convolutional neural network (CNN), where each branch has a backbone CNN (e.g., VGG) to learn feature representations and a keypoint proposal network (KPN) comprised of two convolutional layers. The output feature maps from the backbone's last convolutional layer are fed as input to the KPN. 
KPN produces a score map of dimensions $\frac{H}{s} \times \frac{W}{s} \times D $, where $H$ and $W$ are the input image's height and width respectively, $s$ is the network stride, and $D=2$ is a score whether the particular location is a keypoint or not. Softmax is then applied on $D$ such that each location on the KPN output map has a 2-D probability distribution. This output map can be seen as a keypoint confidence score for a grid-based set of keypoint locations over the 2D image. The density of the keypoint sampling depends on the network stride $s$, which in our case was 16 (i.e. a keypoint proposal every 16 pixels). 
In order to extract a descriptor (dim-2048) for each keypoint, the backbone's feature maps are passed to the region-of-interest (RoI) pooling layer along with a set of bounding boxes each centered at a keypoint location. 
The first pair of branches (A, B) of the network are trained with a triplet loss applied to local features, while a relative pose loss is applied to the keypoint predictions. 
Branch D is trained using a Euclidean loss on the local features and with a consistency loss that attempts to align its keypoint predictions and local representations to those of branch C. Note that branches A, B, and C share their weights, while branch D is a different network. Since branch D receives as input a different modality than the rest and we desire branches C and D to produce the same outputs, their weights during training must be independent. In the following sections, we describe the details of the loss functions and training.

\begin{figure}[t]
   \centering
    \begin{subfigure}[b]{0.22\textwidth}
    	\begin{center}
        \includegraphics[width=1\textwidth]{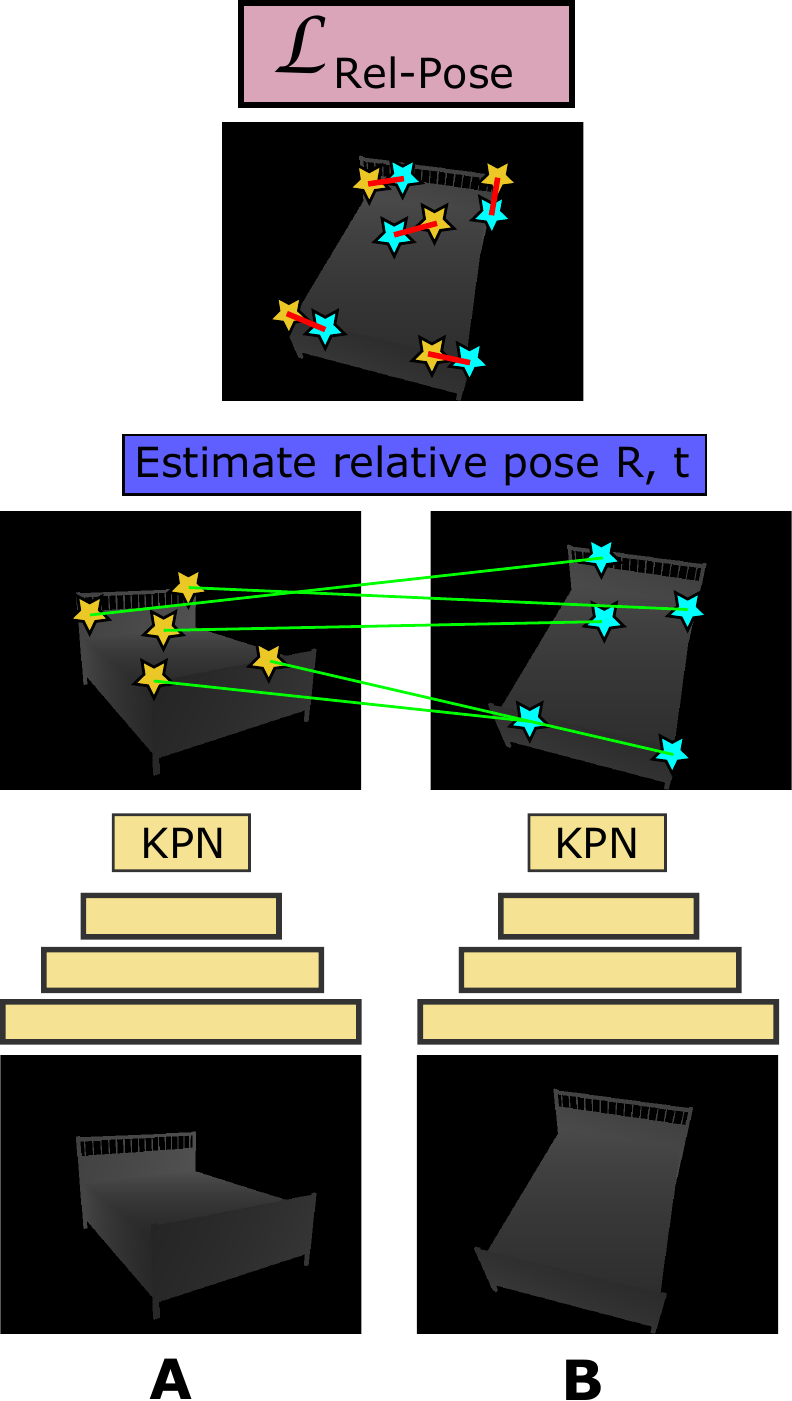}
        \end{center}
        \caption{Relative pose loss.}
        \label{fig:relative_pose}
    \end{subfigure}
        ~
    ~
    \begin{subfigure}[b]{0.22\textwidth}
    	\begin{center}
        \includegraphics[width=1\textwidth]{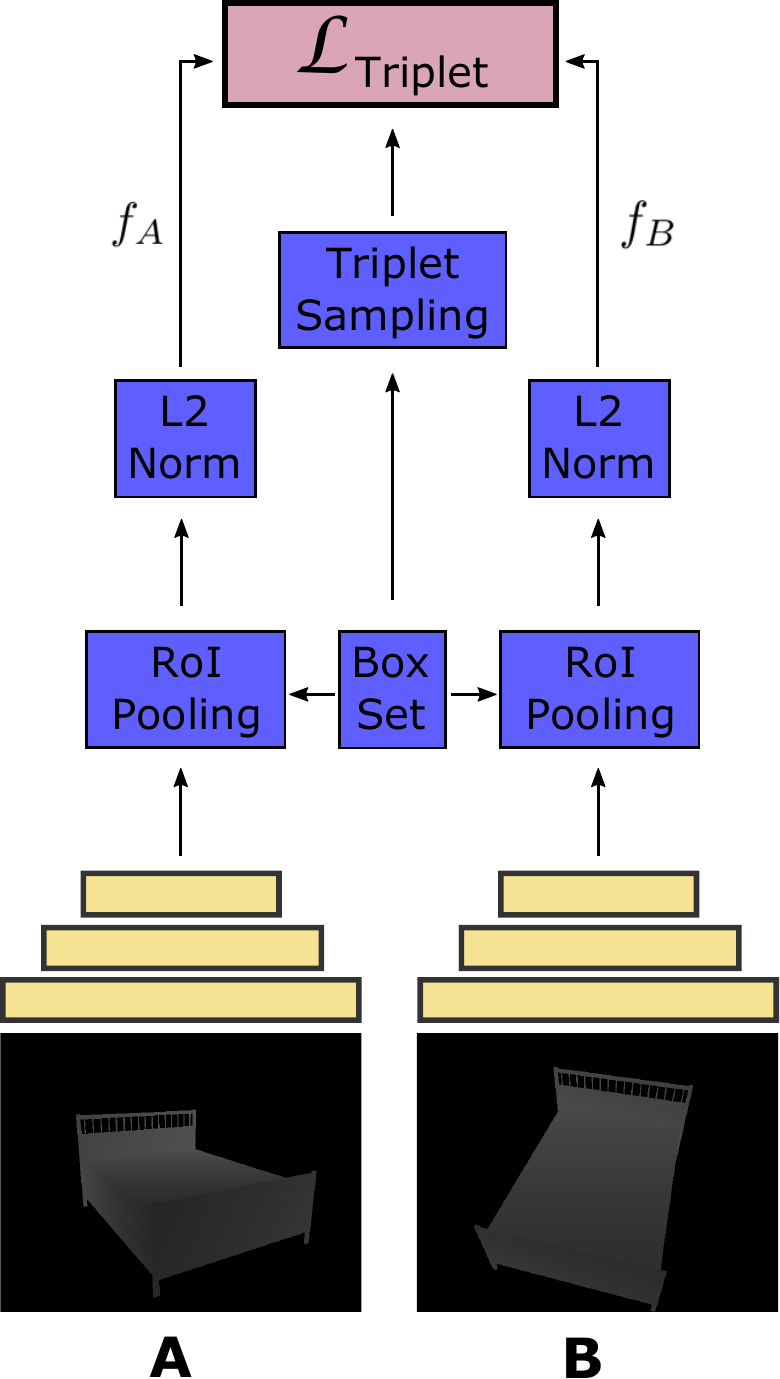}
        \end{center}
        \caption{Triplet loss.}
        \label{fig:triplet}
    \end{subfigure}
    \caption{Relative pose and triplet losses.}
    \vspace{-0.4cm}
    \label{fig:loss_details}
\end{figure}

\begin{figure}[t]
   \centering
    \begin{subfigure}[b]{0.22\textwidth}
    	\begin{center}
        \includegraphics[width=1\textwidth]{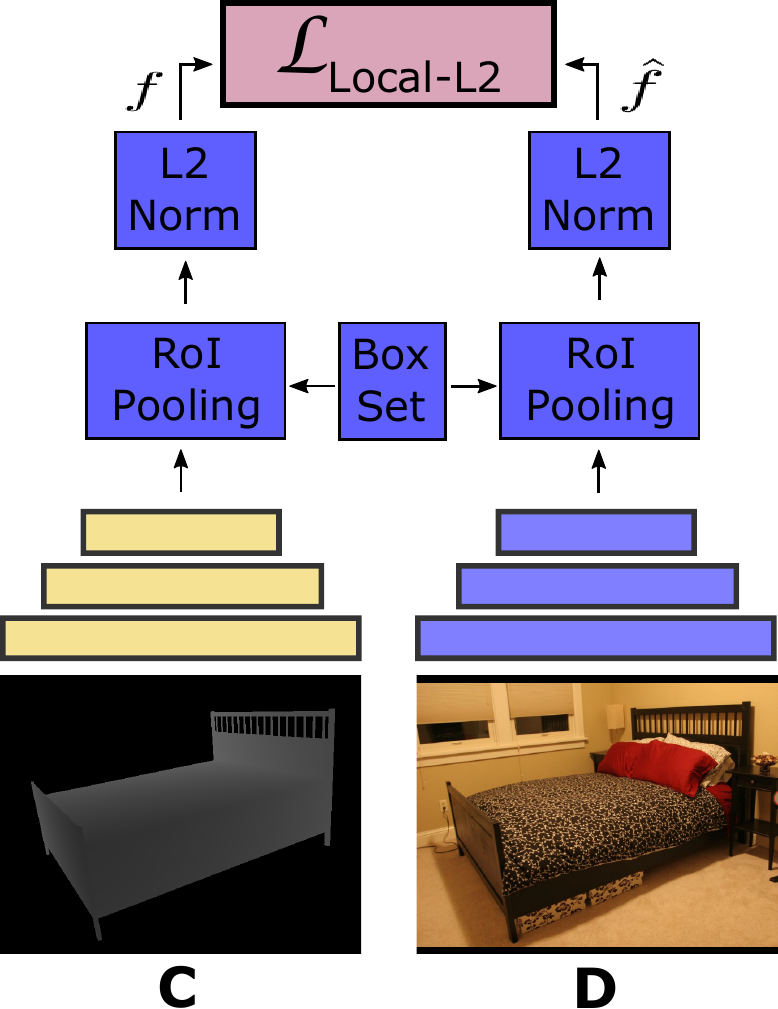}
        \end{center}
        \caption{Local Euclidean loss.}
        \label{fig:local_l2}
    \end{subfigure}
    ~
    ~
   \begin{subfigure}[b]{0.22\textwidth}
    	\begin{center}
        \includegraphics[width=1\textwidth]{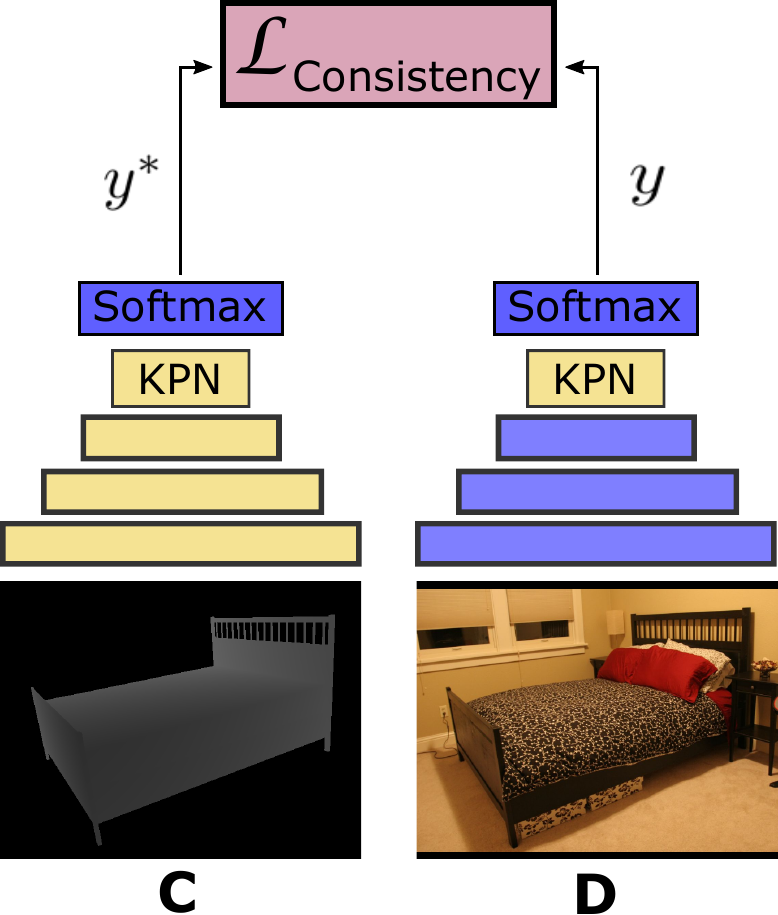}
        \end{center}
        \caption{Keypoint consistency loss.}
        \label{fig:consistency}
    \end{subfigure}
    \caption{Local Euclidean and keypoint consistency losses. }
    \label{fig:loss_details2}
\end{figure}

\subsection{Keypoint Learning by Relative Pose Estimation}
The overall idea behind learning keypoint predictions is to select keypoints that can be used for relative pose estimation
between the input depth images in branches A and B. Specifically, given the two sets of keypoints, we establish correspondences in 3D space, estimate the rotation $R$ and translation $t$, and project the keypoints from depth image A to depth image B. Any misalignment (re-projection error) between the projected keypoints is used to penalize the initial keypoint selections. A pictorial representation of the relative pose objective is shown in Figure~\ref{fig:relative_pose}.

The relative pose objective is formulated as a least squares problem, which finds the rotation $R$ and translation $t$ for which the error of the weighted correspondences is minimal. Formally, for two sets of corresponding points: $P= \{p_1, p_2,\ldots,p_n\}, Q = \{q_1,q_2,\ldots,q_n\}$
we wish to estimate $R$ and $t$ such that:
\begin{equation}
R,t = \argmin_{R \in SO(3), t \in \mathbb{R}^3 } \sum_{i=1}^n w_i ||(Rp_i + t) - q_i||^2
\end{equation}
where $w_i=s_i^A+s_i^B$ is the weight of correspondence $i$ and $s_i^A$ and $s_i^B$ are the predicted keypoint probabilities, as given by KPN followed by a Softmax layer, that belong to correspondence $i$ from branches A and B respectively. Given a set of correspondences and their weights, an SVD-based closed-form solution for estimating $R$ and $t$ that depends on $w$ can be found in~\cite{sorkine2017least}. The idea behind this formulation is that correspondences with high re-projection error should have low weights, therefore a low predicted keypoint score, while correspondences with low re-projection error should have high weights, therefore high predicted keypoint score. With this intuition, we can formulate the relative pose loss as:
\begin{equation}
L_{rel\_pose} = \frac{1}{n} \sum_{i=1}^n w_i g(w_i)
\end{equation}
where $g(w_i) = ||(Rp_i + t)-q_i||^2$. 
Since our objective is to optimize the loss function with respect to estimated keypoint scores, we penalize each keypoint score separately by estimating the gradients for each correspondence and backpropagating them accordingly.

\subsection{Learning Keypoint Descriptors}
In order to match keypoint descriptors across viewpoints, we apply a triplet loss on local features extracted from branches A and B. This involves using the known camera poses of the rendered pairs of depth images and sampling of training keypoint triplets (anchor-positive-negative). Specifically, for a randomly selected keypoint as an anchor from the first image, we find the closest keypoint in 3D from the paired image and use it as a positive, and also select a further away point in 3D to serve as the negative. 
The triplet loss then optimizes the representation such that the feature distance between the anchor and the positive points is smaller than the feature distance between the anchor and the negative points plus a certain margin, and is defined as follows:
\begin{equation}
L_{triplet} = \frac{1}{N} \sum_i^N \max (0, ||f_i^a-f_i^p||^2 - ||f_i^a-f_i^n||^2 + m)
\end{equation}
where $f_i^a$, $f_i^p$, and $f_i^n$ are the local features for the anchor, positive, and negative correspondingly of the $i^{th}$ triplet example and $m$ is the margin. Traditionally, the margin hyper-parameter is manually defined as a constant throughout the training procedure; however, we take advantage of the 3D information and define the margin to be equal to $D_n-D_p$, where $D_n$ is the 3D distance between the anchor and negative, and $D_p$ is the 3D distance between the anchor and positive. Ideally, $D_p$ should be 0, but practically due to the sampling of the keypoints in the image space it is usually a small number close to 0. Essentially this ensures that the learned feature distances are proportional to the 3D distances between the examples and assumes that the features and 3D coordinates are normalized to unit vectors.
Note that the triplet loss only affects the backbone CNN during training and not the KPN. A pictorial representation of the triplet objective is shown in Figure~\ref{fig:triplet}.

\subsection{Cross-modality Representation Learning}\label{subsection:cross-modal}
Finally, we can transfer the learned  features and keypoint proposals from branches (A, B) to branch D, using branch C as a bridge, similar to knowledge distillation techniques~\cite{hinton2015distilling}. To accomplish this, network parameters in branches A, B, and C are shared, and the outputs of branches C and D are compared and penalized according to any misalignment. The core idea is to enforce both the backbone and KPN in branches C and D to generate as similar outputs as possible.  
This objective can be accomplished by means of two key components that are described next.

\noindent
{\bf Local Feature Alignment.}
In order to align local feature representations in branches C and D (see  Figure~\ref{fig:local_l2}),
we consider the predicted keypoints in branch C and compute each keypoint's feature representation, $f_i, i=1,\ldots,k$. 
Keypoint features at corresponding spatial locations from branch D are represented as $\hat{f_i}, i=1,\ldots,k,$.
Formally, we optimize the following objective function:
\begin{equation}
L_{local\_l2} = \frac{1}{k} \sum_{i=1}^k \|\hat{f_i}-f_i\|
\end{equation}
Since we want to align $\hat{f_i}$ with $f_i$, during backpropagation, we fix $f_i$ as ground-truth and backpropagate gradients of $L_{local\_l2}$ only to the appropriate locations in branch D.

\noindent
{\bf Keypoint Consistency.}
Enforcement of the keypoint consistency constraint requires the KPN from branch D to produce the same keypoint predictions as the KPN from branch C. 
It can be achieved using a cross-entropy loss, which is equivalent to a log loss with binary labels: $
L= -\frac{1}{n} \sum_{i=1}^n y_i^* \log y_i $, where $y_i^*$ is the ground-truth label and $y_i$ is the prediction. 
This in our case becomes:
\begin{equation}
L_{consistency} = -\frac{1}{n} \sum_{i=1}^n y_i^C \log y_i^D
\end{equation}
 where $y_i^C$ are the keypoint predictions from branch C, which serve as the ground-truth, and $y_i^D$ are the keypoint predictions from branch D. This loss penalizes any misalignment between the keypoint predictions of the two branches and forces branch D to imitate the outputs of branch C. Figure~\ref{fig:consistency} illustrates inputs to $L_{consistency}$.

\paragraph{Overall objective.}
Our overall training objective is the combination of the losses described above:
\begin{equation}
\begin{split}
L_{all} = \lambda_{1}L_{triplet} + \lambda_{2}L_{rel\_pose} \\ + \lambda_{3}L_{local\_l2} + \lambda_{4}L_{consistency}
\end{split}
\end{equation} 
where each $\lambda$ is the weight for the corresponding loss.

\section{Experiments}
In order to validate our approach, we perform experiments on the Pascal3D+~\cite{xiang2014beyond} dataset and the newly introduced Pix3D~\cite{sun2018pix3d} dataset, which contains 10069 images, 395 CAD models of 9 object categories, and provides precise 3D pose annotations. 
We conduct four key experiments. First, we compare to supervised state-of-the-art methods by training on Pix3D and testing on Pascal3D+ (sec.~\ref{subsection:exp0});
second, we perform an ablation study on Pix3D and evaluate the performance of different parts of our approach (sec.~\ref{subsection:exp1});
third, we test how our model generalizes to new object instances by training only on a subset of provided instances and testing on unseen ones (sec.~\ref{subsection:exp2}); and finally, data from an external dataset, such as NYUv2~\cite{silberman2012indoor} is used to train and test on Pix3D (sec.~\ref{subsection:exp3}). The motivation for the fourth experiment is to demonstrate that our framework can utilize RGB-D pairs from another realistic dataset, where the alignment between the RGB and the depth is provided by the sensor. We use the geodesic distance for evaluation: $\Delta(R_1,R_2) = \frac{||\log(R_1^TR_2)||_F}{\sqrt{2}}$, reporting percentage of predictions within $\frac{\pi}{6}$ of the ground-truth $Acc_{\frac{\pi}{6}}$ and $MedErr$. Additionally, we show the individual accuracy of the three Euler angles, where the distance is the smallest difference between two angles: $\Delta(\theta_1, \theta_2) = \min (2\pi-||\theta_1-\theta_2||, ||\theta_1-\theta_2||)$. For the last metric we also use a threshold of $\frac{\pi}{6}$.

\noindent \textbf{Implementation details.} We use VGGNet as each branch's backbone and start from ImageNet pretrained weights, while KPN is trained from scratch. We set the learning rate to 0.001 and all $\lambda$ weights to 1. In order to regularize the relative pose loss such that it predicts keypoints inside objects, we add a mask term, realized as a multinomial logistic loss. The ground-truth is a binary mask of the object in the rendered depth. This loss is only applied on branches A and B with a smaller weight of 0.25.
Finally, the bounding box dimensions for the RoI layer are set to $32 \times 32$.

\begin{figure*}[t]
\begin{center}
\includegraphics[width=1\linewidth]{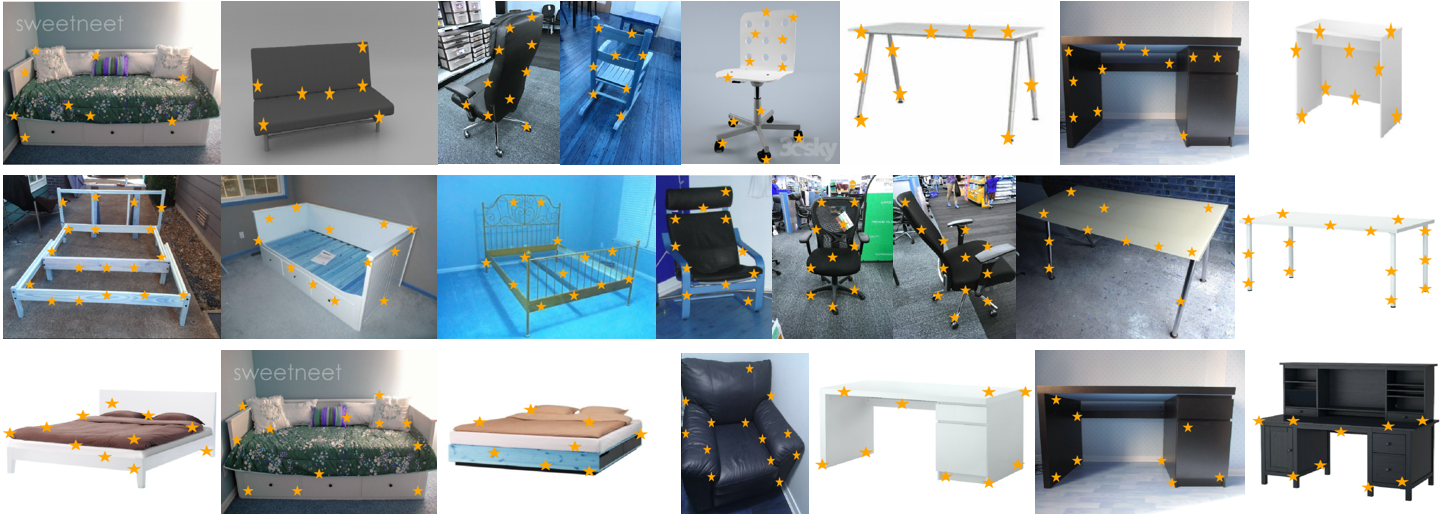}
\end{center}
   \caption{Keypoint prediction examples on test images from the Pix3D dataset. Top, middle, and bottom rows show results from experiments of sections~\ref{subsection:exp1}, \ref{subsection:exp2}, and \ref{subsection:exp3} respectively. Note that we applied non-maximum suppression (NMS) on the keypoint predictions in order to select the highest scoring keypoint from each region.}
\label{fig:keypoints}
\end{figure*}

\begin{table}
\centering
\scalebox{0.75}{
\begin{tabular}{|c|c|c|c|c|}
\hline
Category &\multicolumn{2}{c|}{Chair} &\multicolumn{2}{c|}{Sofa} \\
\hline
Metric & $Acc_{\frac{\pi}{6}} \uparrow$ & $MedErr \downarrow$ & $Acc_{\frac{\pi}{6}} \uparrow$ & $MedErr \downarrow$ \\
\hline\hline
Render for CNN~\cite{su2015render} & 4.3 & 2.1 & 11.6 & 1.2\\
\hline
Vps \& Kps~\cite{tulsiani2015viewpoints} & 10.3 & 1.7 & 23.3 & 1.2\\
\hline
Deep3DBox~\cite{mousavian20173d} & 10.8  & 1.9 & 25.6 & \textbf{1.0} \\ 
\hline 
Proposed & \textbf{13.4} & \textbf{1.6} & \textbf{30.2} & 1.1 \\ 
\hline
\end{tabular}
}
\caption{Comparison with supervised approaches when trained on Pix3D and tested on Pascal3D+. The $MedErr$ is shown in radians.}
\vspace{-0.7cm}
\label{tab:supervised_res}
\end{table}

\begin{table*}
\centering
\scalebox{0.75}{
\begin{tabular}{|c|c|c|c|c|c|c|c|c|c|c|c|c|c|c|c|}
\hline
Category &\multicolumn{5}{c|}{Bed} &\multicolumn{5}{c|}{Chair} & \multicolumn{5}{c|}{Desk} \\
\hline
Metric & Az. & El. & Pl. & $Acc_{\frac{\pi}{6}} \uparrow$ & $MedErr \downarrow$ & Az. & El. & Pl. & $Acc_{\frac{\pi}{6}} \uparrow$ & $MedErr \downarrow$ & Az. & El. & Pl. & $Acc_{\frac{\pi}{6}} \uparrow$ & $MedErr \downarrow$ \\
\cline{1-13}
\hline\hline
Baseline-A & 51.4& 39.1 &35.2 &7.3 & 1.7 & 30.2 & 43.2 & 20.0 & 3.3 & 2.0 & 28.9 & 30.9 &20.4 & 2.6 & 2.2 \\ \hline
Baseline-ZDDA & 48.6 & 50.3 & 41.9 & 21.8 & 1.5 & 35.3 & 48.3 & 26.6 & 11.5 & 1.7 & 24.3 & 23.7 & 21.1 & 3.9 & 2.0\\ \hline
Proposed - joint & 69.8& 51.9 & 58.1 & 31.3 & 1.0 &\textbf{55.3} & \textbf{62.7}& 44.7 &31.1 & \textbf{0.9} &57.2 &48.7 &51.0 & 25.0 & 1.1\\ \hline
Proposed - alternate & \textbf{83.2}& \textbf{67.0} & \textbf{70.4} & \textbf{50.8} & \textbf{0.5} &54.7 & 60.1& \textbf{47.0} &\textbf{31.2} & 1.0 &\textbf{65.1} &\textbf{55.3} &\textbf{58.6} & \textbf{34.9} & \textbf{0.9}\\ \hline

\end{tabular}
}
\caption{Results for azimuth, elevation, in-plane rotation accuracy, $Acc_{\frac{\pi}{6}}$ and $MedErr$ (radians) for the sec~\ref{subsection:exp1} experiment.}
\label{tab:allInst}
\end{table*}

\noindent \textbf{Training data.} All our experiments require a set of quadruplet inputs. For the first two inputs, we first sample from each object's viewsphere and render a view every 15 degrees in azimuth and elevation for three different distances. Then, we sample rendered pairs such that their pose difference is between $\frac{\pi}{12}$ and $\frac{\pi}{3}$. For the last two inputs, we require aligned depth and RGB image pair. 
In order to demonstrate our approach on the Pix3D dataset, we generate these alignments using the dataset's annotations, however, we do not use annotations during training in any other capacity. As we show in sec.~\ref{subsection:exp3}, alternatively the aligned depth and RGB images can be sampled from an existing RGB-D dataset or through hand-alignment~\cite{bansal2016marr}. Note that for each quadruplet, the selection of the first pair of inputs is agnostic to the pose of the object in the last two inputs.

\noindent \textbf{Testing protocol.} For every CAD model instance used in our experiments, we first create a repository of descriptors each assigned to a 3D coordinate. To do so, 20 rendered views are sampled from the viewing sphere of each object, similarly to how the training data are generated, and keypoints are extracted from each view. Note that for this procedure, we use the trained network that corresponds to branch A of our architecture. Then we pass a query RGB image through the network of branch D, generate keypoints and their descriptors and match them to the repository of the corresponding object instance. Finally, the established correspondences are passed to RANSAC and PnP algorithm to estimate the pose of the object. For every keypoint generation step we use the keypoints with the top 100 scores during database creation and top 200 scores for the testing RGB images. When testing on Pix3D, we have defined a test set which contains untruncated and unoccluded examples of all category instances, with 179, 1451, and 152 images in total for \textsl{bed}, \textsl{chair}, and \textsl{desk} category respectively. For Pascal3D+ we follow the provided test sets and make use of the ground-truth bounding boxes.

\begin{table*}
\centering
\scalebox{0.75}{
\begin{tabular}{|c|c|c|c|c|c|c|c|c|c|c|c|c|c|c|c|}
\hline
Category &\multicolumn{5}{c|}{Bed} &\multicolumn{5}{c|}{Chair} & \multicolumn{5}{c|}{Desk} \\
\hline
Metric & Az. & El. & Pl. & $Acc_{\frac{\pi}{6}} \uparrow$ & $MedErr \downarrow$ & Az. & El. & Pl. & $Acc_{\frac{\pi}{6}} \uparrow$ & $MedErr \downarrow$ & Az. & El. & Pl. & $Acc_{\frac{\pi}{6}} \uparrow$ & $MedErr \downarrow$ \\
\cline{1-13}
\hline\hline
Baseline-A & 38.2 & 39.6 & 30.6 & 9.7 & 1.9 & 28.6 & 41.4 & 20.3 & 3.7 & 1.9 & 37.6 & 34.4 & 28.8 & 5.6 & 2.0\\ \hline
Baseline-ZDDA & 29.9 & 39.6 & 22.2 & 4.9 & 2.3 & 30.1 & 44.6 & 21.5 & 7.6 & 1.9 & 36.8 & 43.2 & 30.4 & 13.6 & 1.7\\ \hline
Proposed - joint & 66.7& 50.0 & 62.5 & 29.2 &0.9 &43.7 & 50.4 & 31.3 &15.1 & 1.4 &59.2 &44.0 &41.6 & 13.6 & 1.3\\ \hline
Proposed - alternate & \textbf{75.7}& \textbf{61.1} & \textbf{74.3} & \textbf{45.1} & \textbf{0.6} &\textbf{52.0} & \textbf{57.4} & \textbf{38.0} &\textbf{21.2} & \textbf{1.2} &\textbf{62.4} &\textbf{44.0} &\textbf{53.6} & \textbf{18.4} & \textbf{1.2}\\ \hline
\end{tabular}
}
\caption{Results for azimuth, elevation, in-plane rotation accuracy, $Acc_{\frac{\pi}{6}}$ and $MedErr$ (radians) for the sec.~\ref{subsection:exp2} experiment.}
\label{tab:modelGen}
\end{table*}

\begin{figure*}[!ht]
\begin{center}
\includegraphics[scale=0.4]{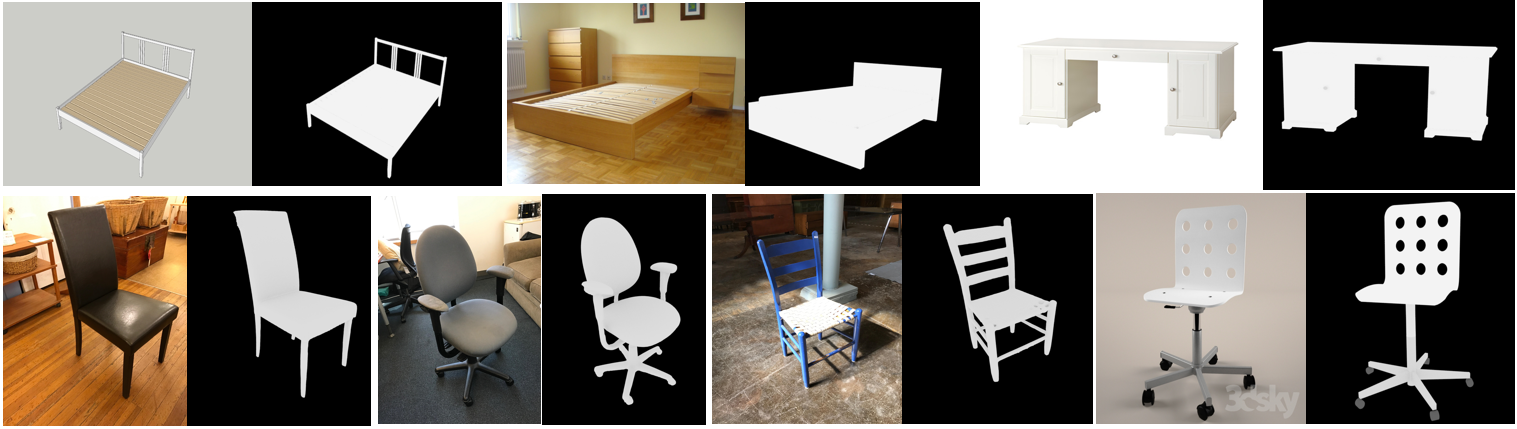}
\end{center}
\vspace{-0.5cm}
   \caption{Illustration of rendered estimated poses on test RGB images from the Pix3D dataset for the sec.~\ref{subsection:exp1} experiment.}
\vspace{-1.5em}
\label{fig:estPoses}
\end{figure*}

\subsection{Comparison with supervised approaches}\label{subsection:exp0}
Given our approach does not use any pose annotations during training, it is challenging to evaluate it against existing state-of-the-art methods, which use pose annotations during training. In addition, our method cannot be trained on Pascal3D+ because it requires paired RGB and depth images, which cannot be generated from the dataset's annotations. 
Therefore, we designed the following experiment for a fair comparison: we train all methods on Pix3D and test on Pascal3D+. 
We compare to the state-of-the-art methods of Deep3DBox~\cite{mousavian20173d}, Render for CNN \cite{su2015render}, and Viewpoints \& Keypoints~\cite{tulsiani2015viewpoints}, all of which require pose annotations for RGB images.
Other approaches, such as Pavlakos \etal~\cite{pavlakos20176}, were considered for comparison but unfortunately they require semantic keypoint annotations during training which Pix3D does not provide. We conduct this evaluation on the common categories between Pix3D and Pascal3D+ (\textsl{chair} and \textsl{sofa}) and report results in Table~\ref{tab:supervised_res}. .

As expected, all approaches generally underperform when applied on a new dataset. Our method demonstrates better generalization and achieves higher $Acc_{\frac{\pi}{6}}$ for both objects, even though it does not explicitly require 3D pose annotations during training. This is due to fundamental conceptual differences between these approaches and ours. These methods formulate viewpoint estimation as a classification problem where a large number of parameters in fully-connected layers are to be learned. This increases the demand for data and annotations and confines the methods mostly to data distributions that were trained on. On the other hand, we exploit CAD models to densely sample from the object's viewsphere, and explicitly bridge the gap between the synthetic data and real images, thereby reducing the demand for annotations. Furthermore, the learned local correspondences allow us more flexibility when it comes to understanding the geometry of unseen objects, as we also demonstrate in sec.~\ref{subsection:exp2}.

\subsection{Ablation study}\label{subsection:exp1}
To understand how each objective of our approach contributes, we have carefully designed a set of baselines, which we train and test on Pix3D, and compare them on the task of pose estimation for the \textsl{bed}, \textsl{chair}, and \textsl{desk} categories. 

\noindent \textbf{Baseline-A.} In order to assess the importance of the cross-modality representation learning (sec.~\ref{subsection:cross-modal}), we learn view-invariant depth representations and depth keypoints and simply use these keypoints and representations during testing. In practice, this corresponds to removing the local euclidean and keypoint consistency losses, and using only the triplet and relative pose losses during training. Consequently this baseline is utilizing only depth data during training, but is applied on RGB images during testing.  

\noindent \textbf{Baseline-ZDDA.} Another baseline would be to only learn RGB-D modality invariant representations, i.e., similar features for RGB and depth images, which can then be used to match RGB images to depth renderings from CAD models. In practice, this would correspond to training our proposed approach with only the local feature alignment objective by sampling all possible keypoint locations. 
This is similar in spirit to and an improved version of ZDDA~\cite{peng2018zero}, a domain adaptation approach that maps RGB and depth modalities to the same point in the latent space. 

\noindent \textbf{Joint and alternate training.} Finally we use all objectives in our approach and investigate two different training strategies. First we try training all objectives jointly in a single optimization session and report this baseline as \textit{Proposed-joint}. Second, we define a three-step alternating training, where we initially optimize using only the triplet and relative pose losses (i.e. branches A, B, C), then we optimize only with the local euclidean and keypoint consistency losses (i.e. branch D), and in the last step all objectives are jointly optimized together. This baseline is reported as \textit{Proposed-alternate}. Note that also experiments in sec.~\ref{subsection:exp0} and ~\ref{subsection:exp3} follow this training paradigm.

\noindent \textbf{Results.} We first show, in Figure~\ref{fig:keypoints} (top row), qualitative keypoint prediction results on test images, where we see keypoint predictions that generally satisfy our intuition of good keypoints. We then adopt the testing protocol described above to report quantitative pose estimation results for test RGB images. 
Performance analysis is shown in Table~\ref{tab:allInst} for the three object categories. As can be noted from the results, our proposed model generally achieves higher accuracy when compared to the baseline approaches. In particular, the improvements over Baseline-A suggests that keypoint and representation modality adaptation enforced in our model is critical. Furthermore, the improvements over Baseline-ZDDA suggests that simply performing modality adaptation for the RGB and depth features is not sufficient, and learning keypoints and view-invariant representations, as is done in our method, is important to achieve good performance. Finally, we observe that alternating training outperforms the joint strategy, demonstrating the importance of learning good keypoints and representations first, before transferring to the RGB modality.

\subsection{Model transferability}\label{subsection:exp2}
In this section, we demonstrate the transfer capability, where the goal is
for a model, trained according to the proposed approach, to generalize well to category instances \textbf{not seen} during training. This is key to practical usability of the approach since we cannot possibly have relevant CAD models of all instances of interest during training. To this end, the baselines introduced in sec.~\ref{subsection:exp1} are re-used with the  following experimental protocol: during training, quadruplets are sampled from a subset of the available instances for each category, and test on RGB images corresponding to all other instances. For instance, for the \textsl{bed} category, we use 10 instances for training and 9 instances for testing. Similarly, for \textsl{chair} and \textsl{desk}, we use 111 and 12 instances respectively for training and the rest for testing. During testing, we use the same protocol as above. We present qualitative keypoint predictions in Figure~\ref{fig:keypoints} (middle row) and report quantitative performance in Table~\ref{tab:modelGen}. As can be seen from the results, our model shows good transferability, providing (a) a similar level of detail in the predicted keypoints as before, (b) improved  accuracy when compared to the baselines, and (c) absolute accuracies that are not too far from those in Table~\ref{tab:allInst}.

\subsection{Framework flexibility}\label{subsection:exp3}
While the results reported above use RGB-D pairs from the Pix3D dataset to train our model, in principle, our approach can be used in conjunction with other datasets that provide aligned RGB-D pairs as well. Such capability will naturally make it easier to train models with our framework, leading to improved framework flexibility. To demonstrate this aspect, we train our model as before, but now for input to branches C and D, we use aligned RGB-D pairs from the NYUv2~\cite{silberman2012indoor} dataset. Since these pairs contain noisy depth images from a real depth sensor, we synthetically apply realistic noise on the clean rendered depth images, used for branches A and B, using DepthSynth~\cite{planche2017depthsynth}. This ensures that branches A, B, and C still receive the same modality as input. 
Note that we do not test on NYUv2, but rather we use it to collect auxiliary training data and perform testing on Pix3D.
Similarly to all other experiments, we do not use any pose annotations for the RGB images as part of training our model and we follow the previous testing protocol. Figure~\ref{fig:keypoints} (bottom row), shows some keypoint prediction results on test data from Pix3D. In Table~\ref{tab:flexEvalAzimuthElevation}, we report quantitative results. We can make several observations- while the numbers are lower than those with the proposed method in Table~\ref{tab:allInst}, which is expected, they are higher than all the baselines reported in Table~\ref{tab:allInst}. Please note that the baselines were trained with alignment from Pix3D, whereas our model here was trained with alignment from NYUv2. These results, along with those in the previous section, show the potential of our approach in learning generalizable models for estimating object poses in RGB images, while not explicitly requiring any pose annotations during training.

\begin{table}
\centering
\scalebox{0.85}{
\begin{tabular}{|c|c|c|c|c|c|}
\hline
Metric & Az. & El. & Pl. & $Acc_{\frac{\pi}{6}} \uparrow$ & $MedErr \downarrow$   \\
\hline\hline
Bed &65.9 & 54.1&44.0 & 24.0 & 1.0 \\ 
\hline 
Chair &44.3&51.0 & 31.0 & 15.2 & 1.6 \\ 
\hline 
Desk &50.0 & 45.4 &31.6 & 7.2 & 1.9  \\ 
\hline
\end{tabular}
}
\caption{Results for sec.~\ref{subsection:exp3} experiment.}
\vspace{-0.7cm}
\label{tab:flexEvalAzimuthElevation}
\end{table}

\section{Conclusions}
We proposed a new framework for 3D object pose estimation in RGB images, which does not require either textured CAD models or 3D pose annotations for RGB images during training. We achieve this by means of a novel end-to-end learning pipeline that guides our model to discover keypoints in rendered depth images optimized for relative pose estimation as well as transfer the keypoints and representations to the RGB modality. Our experiments have demonstrated the effectiveness of the proposed method on unseen testing data compared to supervised approaches, suggesting that it is possible to learn generalizable models without depending on pose annotations.

{\small
\bibliographystyle{ieee}
\bibliography{egbib}
}

\end{document}